\documentclass{svproc}
\usepackage{url}
\usepackage{graphicx}

\begin{document}
\mainmatter              
\title{Evaluating Input Representation for Language Identification in Hindi-English Code Mixed Text}
\titlerunning{Language Identification in Hindi-English Code Mixed Text}  
%
\author{Ramchandra Joshi$^1$ \and
Raviraj Joshi$^2$}
\authorrunning{R. Joshi et al.}
%
\institute{$^1$Department of Computer Engineering, Pune Institute of Computer Technology\\
$^2$Department of Computer Science and Engineering, Indian Institute of Technology Madras\\
\email{\{rbjoshi1309, ravirajoshi\}@gmail.com}}

\maketitle              

\begin{abstract}
Natural language processing (NLP) techniques have become mainstream in the recent decade.  Most of these advances are attributed to the processing of a single language. More recently, with the extensive growth of social media platforms focus has shifted to code-mixed text. The code-mixed text comprises text written in more than one language. People naturally tend to combine local language with global languages like English. To process such texts, current NLP techniques are not sufficient. As a first step, the text is processed to identify the language of the words in the text. In this work, we focus on language identification in code-mixed sentences for Hindi-English mixed text. The task of language identification is formulated as a token classification task. In the supervised setting, each word in the sentence has an associated language label. We evaluate different deep learning models and input representation combinations for this task. Mainly, character, sub-word, and word embeddings are considered in combination with CNN and LSTM based models. We show that sub-word representation along with the LSTM model gives the best results. In general sub-word representations perform significantly better than other input representations. We report the best accuracy of 94.52\% using a single layer LSTM model on the standard SAIL ICON 2017 test set.
\end{abstract}
\section{Introduction}
The automatic processing of text to derive insights has been widely used in the industry. It can be used to process product or movie reviews in order to derive a general sentiment. Other applications include analysis of tweets to derive perception of a brand or thoughts of people on a specific topic. All of these applications can be boiled down to classification or summarization tasks. The state of the art NLP techniques performs very well on these tasks for single language texts. However, they may not perform well on code-mixed text due to the unavailability of enough labeled data. The code-mixed text has become relevant these days because of different social media platforms \cite{gamback2016comparing} \cite{barman2014code}. Most of these platforms prefer English as the preferred medium of communication. In a multi-lingual country like India people tend to mix local language with English while using social media platforms. This is because people are more comfortable in local languages. It is natural to describe local terms or entities in local languages which result in code-mixing. Code-mixing essentially allows us to borrow terms from different languages thus aiding ease of communication. A local touch can be given to the movie reviews, product reviews, and comments by adding some details in the local language. All of these factors have led to a rise in the popularity of code-mixed text \cite{khanuja2020gluecos}. In order to understand such code-mixed text, it is important to identify the language used in different parts of text followed by language-specific processing \cite{das2015code}. In this work, we present different approaches for language identification in code-mixed text. In the code-mixed text, languages can be interleaved in different forms. One form of code-mixing is represented in this example, \textit{"this is not union budget, ye to aam admi ka budget hai"} with language tagged as "eng eng eng eng eng hin hin hin hin hin hin hin". Another form can be seen as \textit{"maine aaj WhatsApp and Facebook uninstall kiya h"} tagged as "hin hin eng eng eng eng hin hin". It is challenging to determine the language of individual words as the same word can be used in both Hindi and Engish depending on the context. For example English words like \textit{"are"} and \textit{"maze"} can also be used in Hindi as \textit{"are mai ghar jaa raha hu"}, and \textit{"appke to maze hai"}. To make it more challenging the social media text is normally noisy where words are written in different ways just to emphasize them. For example the in word \textit{"good"} the letter \textit{'o'} can be repeated multiple times to get different variations like \textit{"the movie was gooood"}. This makes it important to consider different input representations. The sentence can be processed word by word or character by character. The more recent form of representation is sub-word where a word is split into logical sub-word units \cite{kudo2018subword}. The character and sub-word based representations are more agnostic to noisy text variations as compared to word representation which will treat each variation as a separate word. The focus of our work is to evaluate the performance of these input representations. This is the first work to explore sub-word based representations for Hi-En language identification. The task is to determine the language of each word in the sentence. The task is formulated as a token classification task. The deep learning models based on convolutional neural networks (CNNs) and long short term memory (LSTM) networks are the most popular techniques used for token classification. We use these simple models in combination with different input representations to evaluate their effectiveness. These models are often used with a conditional random field (CRF) to improve the performance. However, the work is restricted to simple architectures with a focus on input representation. We show that sub-word based representation coupled with these simple models perform better than other complex architectures reported in the literature. Simple architectures are also favorable as it reduces runtime speed and complexity. The language identification module should be fast and efficient as it will often be followed by other NLP modules. With this perspective, we experiment with single-layer CNN and LSTM models. The experiments show that these architectures are sufficient to reach desired accuracy levels. The main contributions of this work are:
\begin{itemize}
    \item The effectiveness of character, sub-word, and word-based representations are evaluated for the task of Hindi-English language identification.
    \item The combinations of popular model architectures and input representations are also compared.
\end{itemize}

\section{Related Work}
In this section, we review some of the deep learning-based approaches used for code mixed language identification. Simple feed-forward neural networks utilizing character n-gram and lexicon features have shown to produce a good performance for this task \cite{zhang2018fast}. Although the task can be performed at the word level, it is a common practice to use neighboring words contextual information to aid the classification process \cite{samih2016multilingual}. The word vectors can be directly passed to a bi-directional LSTM to encode the contextual information. Alternatively, both word vectors and character-based word vectors can be provided to the LSTM. The character-based word representations can be generated using another CNN or LSTM \cite{mandal2018language}. Multi-channel CNNs are commonly used to capture such character-based word representations \cite{kim2014convolutional}. The primary motive behind using character-based representation is to avoid the out of vocabulary problem. It also helps in better classification of words that have very low representation in the training data. There have been few works related to code mixed Hindi-English languages applying similar concepts \cite{veena2018character}. The Bengali-English code mixed text has also been equally explored in literature \cite{chanda2016unraveling}. Other Indian languages like Telugu and the Assamese have also be analyzed from the code-mixing perspective \cite{gundapu2018word} \cite{sarmaidentifying}. We have limited ourselves to Hindi-English text because of the lack of standardized and meaningful data sets in other languages.    
We study the usage of plain word embedding and its combination with character-based word embedding. Sub-word based representations are also evaluated to highlight its importance in identifying code-mixed languages.

\section{Dataset}
The ICON 2017 code mixed sentiment analysis data set is used for evaluation \cite{patra2018sentiment}. The data set also has each and every word tagged with its corresponding language. The sentiment tags are ignored and only language information is used for the experiments. The code mixed text in this data set was extracted from twitter and manually annotated for sentiment and language. There are a total of 12936 training sentences and 5525 test sentences. The split is predefined and the results reported are using the same split. A small validation set comprising of 10\% of train sentences is using for hyper-parameter tuning of the models. There are around 80k Hindi and English tokens individually in the train data. In the test data, there are around 30k tokens in each class.

\begin{figure}[h]
    \centering
    \includegraphics[scale=0.23]{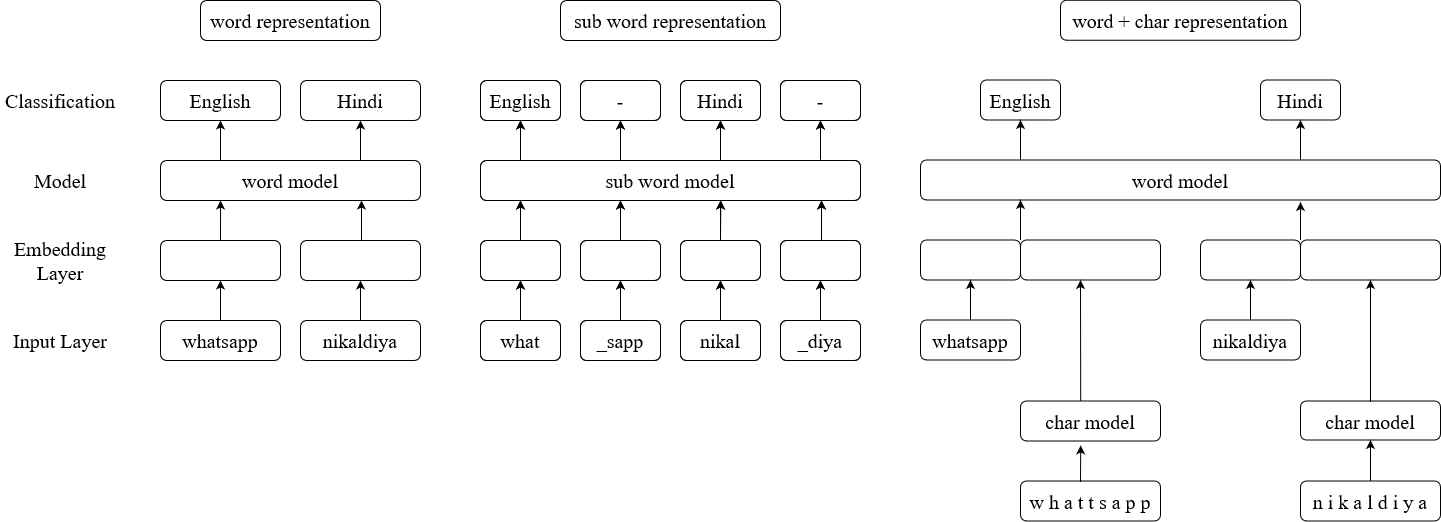}
    \caption{Input Representations}
    \label{fig:word_embed}
\end{figure}

\section{Input Representations}
The input representations evaluated in this work are characters, sub-words, and words. The structure of each representation is shown in Figure \ref{fig:word_embed}.
    \subsection{Word representation}
    The word embeddings are the default representations used for processing text. In this approach, each word is considered as a token and the sentence can be seen as a series of tokens. Each token is represented using a 300-dimensional vector also known as the word vector. This distributed representation allows us to capture the semantic relationship between individual words. The sentence can now be seen as a series of word vectors that are processed using CNN or LSTM to get their contextual representations. The output of CNN or LSTM is again a series of vectors but these vectors now encode the contextual information as opposed to word vectors. This contextual vector is passed through dense layers to get final predictions. So, each word corresponds to a time step, and for each time step its corresponding label in terms of language id is predicted.
    \subsection{Character representation} 
    Another approach is to use character-based representations to augment the word vectors. The idea is to use another shallow network to process each word character by character. The series of characters can now be seen as a time series that is passed through CNN to get contextual character representation. The representations are max-pooled over time to get corresponding word representations. These character-based word representations are then concatenated with token-based word representations discussed earlier and then processed using neural networks. Simply using character-based word representations performs similar to the first word-based approach at a cost of increased complexity. So we only report results for model where character representations are used in conjunction with word embeddings. The usage of character-based word embeddings mitigates the out of vocabulary problem to some extent. The token-based word representations for unknown words will be mapped to a single unk token and the neural network will have to rely on contextual information to make predictions. Whereas the character-based word representations will always give us meaningful representations since the number of characters is fixed and known.
    \subsection{Sub-word representation}
    The final representation strategy explored here is the sub-word embeddings \cite{kudo2018sentencepiece}. This can be seen as an intermediate form with granularity somewhere between words and characters. Some of the possible sub words for word "hello" are "hell/o", "he/llo", and "he/l/l/o". This type of representation is very useful to mitigate the open vocabulary problem. The exact sub word split is determined by the statistical character n-gram properties of the training corpus. We train a uni-gram based subword model using Google sentence piece tokenizer \cite{kudo2018subword}. The subword vocab size is set to 12k. This subword model is used to split each word into constituent sub words. The first subword of each word is assigned the parent language label. The subsequent subwords are assigned a dummy label. The problem is again formulated as a token classification problem where we are only concerned about the label of the first sub-word token of each word. However, during cross-entropy training, all the tokens contribute to the loss. The masking of loss from dummy labels is not explored in this work.  

\section{Model Architecture}
\begin{itemize}
    \item \textbf{CNN}: This is a basic CNN model based on 1D convolutions. The word or sub-word embeddings of 300 dimensions are passed through a single 1D convolution. The kernel size is 4 and the number of filters is 64. This is followed by two dense layers of size 100 and 1. The output of the individual time step is subjected to a dense layer so as to have a prediction for each word. The convolutional layer and dense layers are followed by relu and sigmoid activation functions respectively in all the models described here. Adam is used as an optimizer. The binary cross-entropy is used as the loss function as there are two output classes. The optimizer and loss function are common across all the models.
    \item \textbf{Multi-CNN}: In this model, three parallel 1D convolutions are applied on the word or sub-word embeddings. The filter sizes are 2, 3, and 4 with 64 filters each. The output of these convolutions is concatenated. This is followed by two dense layers of size 100 and 1. 
    \item \textbf{LSTM}: This is a basic LSTM model. A single Bi-LSTM with 300 hidden units is used. The word or sub-word embeddings of 300 dimensions are passed through this layer. The output at each time step is subjected to two dense layers of size 100 and 1. A dropout of 0.4 is used in the recurrent layer.  
    \item \textbf{CNN+LSTM}: This combines the basic CNN and LSTM models sequentially. The 1D CNN as described above is followed by the Bi-LSTM layer. The output is then subjected to two dense layers.  
    \item \textbf{CharCNN+LSTM}: The three parallel convolutions as described in the Multi-CNN network is used to process characters. The output of each convolutional layer is max-pooled over time and concatenated to produce word embedding of size 192 dimension. The time axis corresponds to the number of characters in the word. The word embedding generated by this multi-cnn network is then concatenated with 300 dimension learnable word embeddings as used in previous models. The concatenated representations are passed through a single Bi-LSTM layer and dense layers. The post embedding setup is the same as the basic LSTM model described above.  
\end{itemize}

 
 
 
 

\begin{table}
\caption{Classification metrics of different models and input combinations}\label{tab1}
\label{tab:1}       
%
%
\begin{tabular}{p{2.6cm}p{2cm}p{1cm}p{1.5cm}p{1.5cm}p{1.5cm}p{1cm}}
\hline\noalign{\smallskip}
\textbf{Model} & \textbf{Input} & \textbf{lang} & \textbf{precision} & \textbf{recall} & \textbf{f1-score} & \textbf{acc}  \\
\noalign{\smallskip}\hline\noalign{\smallskip}
CNN & word & en & 90.42  & 92.15  & 91.27 & 91.49 \\
 &  & hi  & 90.52  & 90.87 & 91.69  & \\
 & sub-word & en & 95.21  & 94.97  & 95.09 & 93.86 \\
 &  & hi  & 91.62  & 92.01 & 91.81 & \\
\noalign{\smallskip}\hline\noalign{\smallskip}

Multi-CNN & word & en & 91.68  & 91.00  & 91.34 & 91.66 \\
 &  & hi  & 91.64  & 92.28 & 91.96 & \\
 & sub-word & en & 95.60  & 94.99  & 95.30 & 94.13 \\
 &  & hi  & 91.71  & 92.69 & 92.20 & \\
\noalign{\smallskip}\hline\noalign{\smallskip}

LSTM & word & en & 91.25  & 91.40  & 91.33 & 91.61 \\
 &  & hi  & 91.95  & 91.81 & 91.88 & \\
 & sub-word & en & \textbf{95.15}  & \textbf{96.13}  & \textbf{95.64} & \textbf{94.52}\\
 &  & hi  & \textbf{93.41}  & \textbf{91.81} & \textbf{92.60} & \\
\noalign{\smallskip}\hline\noalign{\smallskip}

CNN+LSTM & word & en & 91.36  & 91.61  & 91.48 & 91.76 \\
 &  & hi  & 92.13  & 91.90 & 92.02 & \\
 & sub-word & en & 95.82  & 95.17  & 95.50 & 94.39 \\
 &  & hi  & 92.01  & 93.06 & 92.54 & \\
\noalign{\smallskip}\hline\noalign{\smallskip}

CharCNN+LSTM & char+word & en & 91.94  & 93.06  & 92.49 & 92.71 \\
 &  & hi  & 93.44  & 92.39 & 92.91 & \\
\noalign{\smallskip}\hline\noalign{\smallskip}

\end{tabular}
\end{table}

\section{Results and Discussion}
A combination of models and input representations is evaluated on the ICON 2017 data set. The word and sub-word representations are used as input to CNN and LSTM based models. The models used are simple CNN, simple LSTM, Multi-CNN, and CNN+LSTM. The character representations are used with the CharCNN+LSTM model. In this model, characters are processed using Multi-CNN and passed to LSTM along with word representation. This model architecture was chosen as it performs better than other variations of character embedding based models. Their variations can have either CNN or LSTM for processing characters and the main network can again be based on either CNN or LSTM. The metrics used for evaluation are precision, recall, f1-score, and overall accuracy. Table \ref{tab1} shows the results for model and input combinations. The LSTM model utilizing subword embeddings performs the best. Augmentation of character-based word embeddings with vanilla word embeddings boosts the performance of word-based models. Thus performance-wise word $<$ char + word $<$ sub-word relationship holds. From the model perspective CNN $<$ Multi-CNN $<=$ LSTM when the only word or sub-word embeddings are considered. The CNN+LSTM gives the best score when word embeddings are given as input to the models. In general sub-word based models are significantly better as compared to models utilizing word and character representations. This is primarily because sub words not only handle unknown words but also go well with the noisy data.       
%

\section{Conclusion}
In this work, we study different deep learning approaches for language identification in Hindi-English code mixed text. The problem can be seen as a token classification problem. A combination of different models and input representations are compared for their effectiveness. The models include simple CNN, simple LSTM, multi-channel CNN, CNN+LSTM, and charCNN+LSTM. The input representations used are characters, words, and subwords. We show that sub-word based models are superior as compared to word and character-based approaches. The sub word representation when coupled with simple LSTM performs the best. Just changing the input representation helps basic models achieve high performance. The sub word representation helps tackle the out of vocabulary problem and at the same time handles noisy data well. These attributes are very analogous with the data set explored in this work.  
%
%
\bibliographystyle{splncs03}
\bibliography{main}
\end{document}